\documentclass{article}

    \usepackage[final]{neurips_2024}

\usepackage[utf8]{inputenc} 
\usepackage[T1]{fontenc}    
\usepackage{hyperref}       
\usepackage{url}            
\usepackage{booktabs}       
\usepackage{amsfonts}       
\usepackage{nicefrac}       
\usepackage{microtype}      
\usepackage[table]{xcolor}         
\usepackage{graphicx}
\usepackage{multicol}
\usepackage{multirow}
\usepackage{makecell}
\usepackage{listings}
\usepackage{bm}
\usepackage{subfig}
\usepackage{amsmath}
\usepackage{amsfonts,amssymb}
\usepackage{wrapfig}
\usepackage{appendix}
\usepackage{colortbl}
\usepackage{adjustbox}

\NewDocumentCommand{\xt}{ mO{} }{\textcolor{red}{\textsuperscript{\textit{XT}}\textsf{\textbf{\small[#1]}}}}

\NewDocumentCommand{\qz}{ mO{} }{\textcolor{blue}
{\textsuperscript{\textit{QZ}}\textsf{\textbf{\small[#1]}}}}
\title{Visual Anchors Are Strong Information Aggregators\\For Multimodal Large Language Model}

\author{Haogeng Liu${}^{1,2}$, Quanzeng You${}^{3}$, Xiaotian Han${}^{3}$, \textbf{Yongfei Liu}${}^{3}$,\\ \textbf{Huaibo Huang}${}^{1,2}$\thanks{Corresponding author},
\textbf{Ran He}${}^{1,2}$, \textbf{Hongxia Yang}\\
${}^{1}$MAIS \& NLPR, Institute of Automation, Chinese Academy of Sciences \\
${}^{2}$School of Artifcial Intelligence, University of Chinese Academy of Sciences \\
${}^{3}$ByteDance, Inc \\
{\tt liuhaogeng22@mails.ucas.ac.cn, \tt huaibo.huang@cripac.ia.ac.cn}\\
}

\begin{document}

\maketitle

\begin{abstract}
In the realm of Multimodal Large Language Models (MLLMs), vision-language connector plays a crucial role to link the pre-trained vision encoders with Large Language Models (LLMs).
Despite its importance, the vision-language connector has been relatively less explored. In this study, we aim to propose a strong vision-language connector that enables MLLMs to achieve high accuracy while maintain low computation cost. We first reveal the existence of the visual anchors in Vision Transformer and propose a cost-effective search algorithm to  extract them. Building on these findings, we introduce the Anchor Former (AcFormer), a novel vision-language connector designed to leverage the rich prior knowledge obtained from these visual anchors during pretraining, guiding the aggregation of information. 
Through extensive experimentation, we demonstrate that the proposed method significantly reduces computational costs by nearly two-thirds compared with baseline, while simultaneously outperforming baseline methods. This highlights the effectiveness and efficiency of AcFormer. Codes are available at \href{https://github.com/liuhaogeng/Anchor-Former}{https://github.com/liuhaogeng/Anchor-Former}.
\end{abstract}
\begin{figure}[h]
  \centering
  \includegraphics[width=13cm]{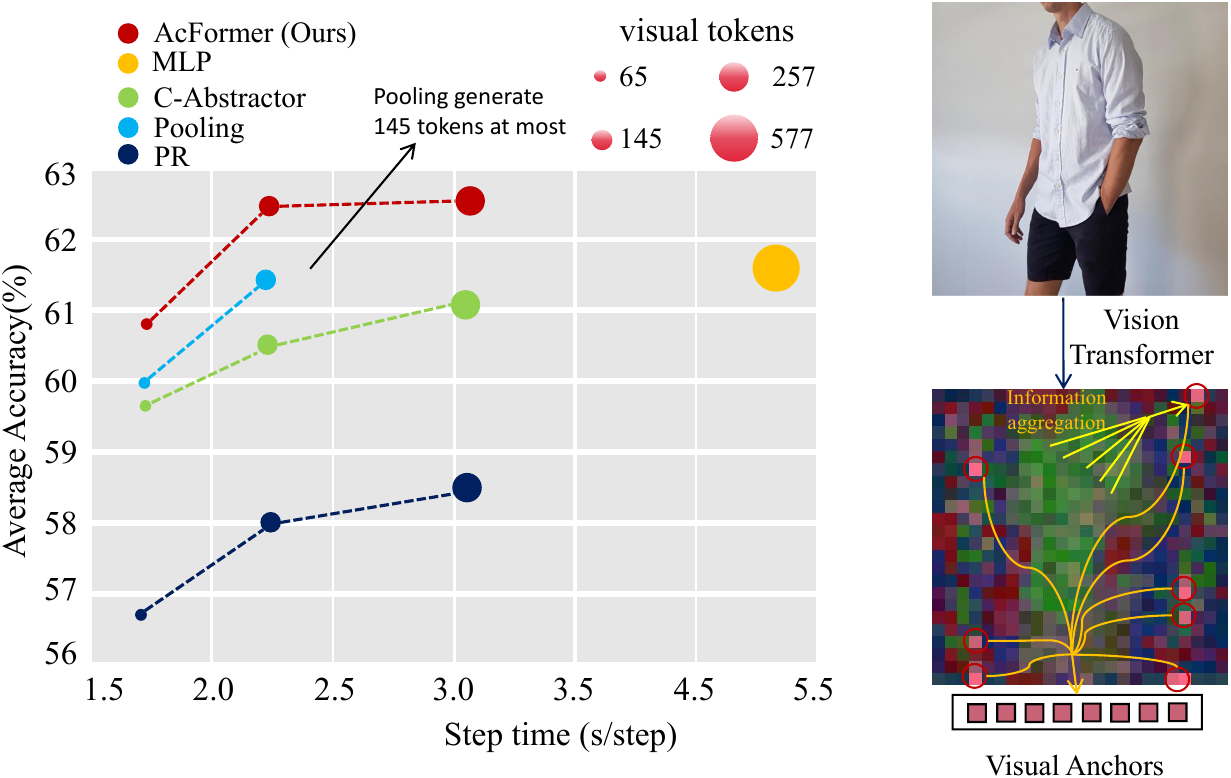}
  \caption{Comparison of the \textbf{average normalized accuracy} (MMB, TextVQA, GQA). PR means Perceiver Resampler, which utilize the learnable query as information aggregator. Our method achieves \textbf{highest accuracy} comparing with the others while maintaining \textbf{high training speed}.}
  \label{tongji}
\end{figure}
\section{Introduction}
Multimodal Large Language Models (MLLMs) have emerged as a focal point within contemporary research discourse\cite{yin2024survey}. Prominently showcased by seminal works such as LLaVA \cite{liu2023improvedllava, liu2024visual}, BLIP-2 \cite{pmlr-v202-li23q}, Qwen-VL \cite{bai2023qwenvl}, and Flamingo \cite{NEURIPS2022_960a172b}, these models exhibit exceptional efficacy across a broad spectrum of tasks, spanning from nuanced image description \cite{li2022blip, yu2022coca} to complex visual reasoning. Their versatility transcends conventional boundaries, finding practical application in quotidian scenarios such as smartphone interface design \cite{hong2023cogagent} and consequential real-world decision-making processes \cite{driess2023palm}. This advancement is attributed to the availability of pre-trained Large Language Models (LLMs) \cite{touvron2023llama} and vision encoders \cite{sun2023eva}. By utilizing these pre-trained components and introducing a connecting layer between them, it is possible to construct robust MLLMs with only training the lightweight vision-language connector. This enables the development of MLLMs with the capacity to process visual inputs while retaining the linguistic prowess characteristic of LLMs. These enhanced MLLMs exhibit proficiency across various tasks, including narrative generation, code composition, and addressing complex queries \cite{wang2023cogvlm}. 

In the construction of above MLLMs, the vision-language connector plays a pivotal role. 
A fundamental approach, as demonstrated in LLaVA \cite{liu2024visual}, employs a linear projection layer as the connector. In the enhanced version, LLaVA-1.5 \cite{liu2023improvedllava}, the linear projection layer is expanded into a multilayer perceptron, enhancing the model's efficacy. Despite achieving notable performance, the large number of visual tokens, which extends the time required for computing attention and other processes \cite{chen2024image}, indicating potential for optimization to decrease the computation cost. Prior efforts have sought to address this concern. For instance, BLIP-2 \cite{pmlr-v202-li23q} introduces Q-Former and Flamingo \cite{NEURIPS2022_960a172b} proposes Perceiver Resampler, both leveraging learnable queries as visual information aggregators. 
This mechanism utilizes the cross-attention between learnable queries and the outputs of visual encoders to effectively reduce the length of the visual sequence, thereby lowering the computation cost. Similarly, Qwen-VL \cite{bai2023qwenvl} adopts a comparable structure but eliminates self-attention among the learnable queries. While these vision-language connectors substantially improve efficiency compared to naive linear projectors, they also exhibit a notable decrease in accuracy, as detailed in \cite{cha2023honeybee}.

The primary computational bottleneck in MLLMs is the LLM component, where computational costs escalate significantly with the length of the input sequence, including both visual and textual tokens. To mitigate the computational cost, a straightforward approach is to reduce the number of input visual tokens with vision-language connector. A commonly adopted method is attention pooling, which offers greater flexibility than traditional pooling techniques. This method focuses on aggregating information within the visual tokens, necessitating aggregators with high information-gathering capabilities. Current attention pooling methods typically use randomly initialized learnable queries as information aggregators. We identify two main drawbacks with this approach: (1) The queries are randomly initialized without prior knowledge for aggregating visual information. They necessitates training on massive datasets (hundreds of millions) to be effective, as showed in Table \ref{ablation_scale} (2) The queries are fixed and invariant to different input images, potentially leading to significant information loss and low specificity for uncommon inputs, as illustrated in \cite{huang2024dq, ren2024grounding}.

To address the aforementioned issue, we propose Anchor Former (AcFormer), a novel vision-language connector that enhances both the accuracy and efficiency of MLLMs compared with baselines. In order to build AcFormer, we identify more effective information aggregators by analyzing the visual features obtained from a pre-trained vision encoder from two perspectives: the feature map and the attention map. Our analysis reveals the presence of ``visual anchors''  within the visual tokens. Fundamentally, the transformers in the neural network aggregate information related to these visual anchors, central to the transformation process. Moreover, the positions of these anchors vary across different images. Despite this variability, we can effectively identify them using the attention matrix to carry out the cost-effective progressively search algorithm. With these observations, we propose a Anchor Selector, an important part of our AcFormer. The Anchor Selector utilize the progressive search algorithm with respect to the attention map. By this way, it effectively extracts visual anchors from the visual tokens generated by the Vision Transformer. And with the visual anchors, AcFormer utilize the naive cross attention to aggregate visual information for generating dense and complete visual representation.

In summary, our contributions can be summarized as follows:
\begin{itemize}
    \item We reveal the existence of visual anchors within the visual tokens generated by pre-trained Vision Transformer, and subsequently propose cost-effective Anchor Selector to effectively extract these visual anchors.
    \item We propose the Anchor Former (AcFormer), a novel vision-language connector designed to improve the accuracy and efficiency of Multimodal Large Language Models (MLLMs) by leveraging the rich prior of information aggregation within visual anchors.
    \item We conduct comprehensive experiments across various vision-language tasks to empirically validate the efficacy of AcFormer.
\end{itemize}

\section{Related Work}
\subsection{Multimodal Large Language Models}
The development of MLLMs has become financially viable due to the utilization of pre-trained vision encoders \cite{sun2023eva, radford2021learning, dosovitskiy2020image} and Large Language Models (LLMs) \cite{touvron2023llama, team2024gemma, zheng2023judging}. Spearheaded by initiatives like Flamingo \cite{NEURIPS2022_960a172b} and BLIP-2 \cite{li2023blip}, the field has witnessed significant advancements \cite{liu2024visual, sun2023generative, zhu2023minigpt4, awadalla2023openflamingo, zhang2023llama, wang2023cogvlm, hong2023cogagent}. 
Studies like LLaVA and MiniGPT-4 have introduced methodologies such as visual instruction tuning, enabling robust MLLMs capable of understanding human instructions. 
Additionally, efforts such as Emu and LaVIT \cite{jin2023unified} have proposed unified frameworks for generation and comprehension, integrating visual decoders and consolidating the training loss of visual and textual inputs. The progression of MLLMs has been supported by the availability of extensive visual-language training datasets \cite{chen2023sharegpt4v, liu2023improvedllava}. Innovations like Sphinx \cite{lin2023sphinx} and Monkey \cite{li2023monkey} have facilitated high-resolution image processing through techniques like sub-image cropping, thus advancing open-source MLLMs. Moreover, MobileVLM \cite{chu2023mobilevlm} has introduced a compact vision language model suitable for deployment on mobile devices.

\subsection{Vision-Language Connectors}

Vision-language connectors typically employ either direct linear projection (LLaVA) or an information aggregation module, such as Flamingo, BLIP-2, and C-Abstractor, followed by linear projection. LLaVA, utilizing a simple multi-layer perceptron, effectively aligns visual features with the embedding space of Large Language Models (LLMs), but it suffers from high computational costs due to redundant input visual tokens, as highlighted by \cite{chen2024image}. BLIP-2 uses the Q-Former to aggregate visual information and establish robust baselines, while Flamingo employs the Perceiver Resampler. Both architectures leverage the cross-attention mechanism to aggregate visual information into learnable queries. However, these approaches require extensive data for training and may have limitations in tasks requiring fine-grained visual perception due to the constrained nature of the learned query's ability to capture all visual patterns. In contrast, Honeybee \cite{cha2023honeybee} proposes the C-Abstractor and D-Abstractor to address these challenges by introducing spatial priors into the feature representation. Our proposed method, AcFormer, consists of three parts: Anchor Selector, Information Aggregation Module and Linear Projection. While Flamingo and BLIP-2 use learnable queries for information aggregation, C-Abstractor applies a convolution network directly. Our method utilizes visual anchors as information aggregators, generating dense and complete visual representations for input images.
\section{Methods}
\subsection{Preliminaries}
For MLLMs, their visual encoders are typically off-the-shelf pre-trained Vision Transformers (CLIP). Given input images $\mathbf{I} \in \mathbb{R}^{B \times C \times H \times W}$ and a Vision Transformer denoted as $F(\cdot)$, the vision feature $\mathbf{R_{v}}$ is obtained as follows:
\begin{align}
\mathbf{R}_{v} = F(\mathbf{I}) \in \mathbb{R}^{B \times N \times D}.
\end{align}
Here, $C$, $H$, and $W$ represent the channel, height, and width of the input image, respectively, while $N$ and $D$ denote the number and dimension of the image tokens. 
There is one additional token, the [CLS] token added for the global representation of the image in contrastive learning. After obtaining the feature map, two common methods are used to combine it with LLM. 

The first method utilizes gated cross-attention for modality fusion. Assuming the corresponding input instruction with the image is $\mathbf{T} \in \mathbb{R}^{B \times N_t \times D_t}$, the computation can be expressed as:
\begin{align}
\mathbf{T}_{h} = G(\text{query}=\mathbf{T}, \text{key}=\mathbf{R}_{v}, \text{value}=\mathbf{R}_{v}),
\end{align}
where $\mathbf{T_{h}}$ represents the hidden states of the LLM and $G(\cdot)$ denotes the gated cross-attention layer.

The other approach involves converting the visual tokens into soft embeddings nd concatenating them with the text embeddings as the input of the LLM. Suppose the vision-language connector is V-L-Connector, 
\begin{align}
    \mathbf{LM}_{in} = \text{Concat}(\text{Proj}(\text{V-L-Connector}(\mathbf{R}_v)), \mathbf{T}),
\end{align}
Where Proj means the Linear Projection and $\mathbf{T}$ means the text embedding. $\mathbf{LM}_{in}$ represents the input embeddings of the LLM.

\subsection{Visual Anchors}
\begin{figure}
  \centering
  \includegraphics[width=14cm]{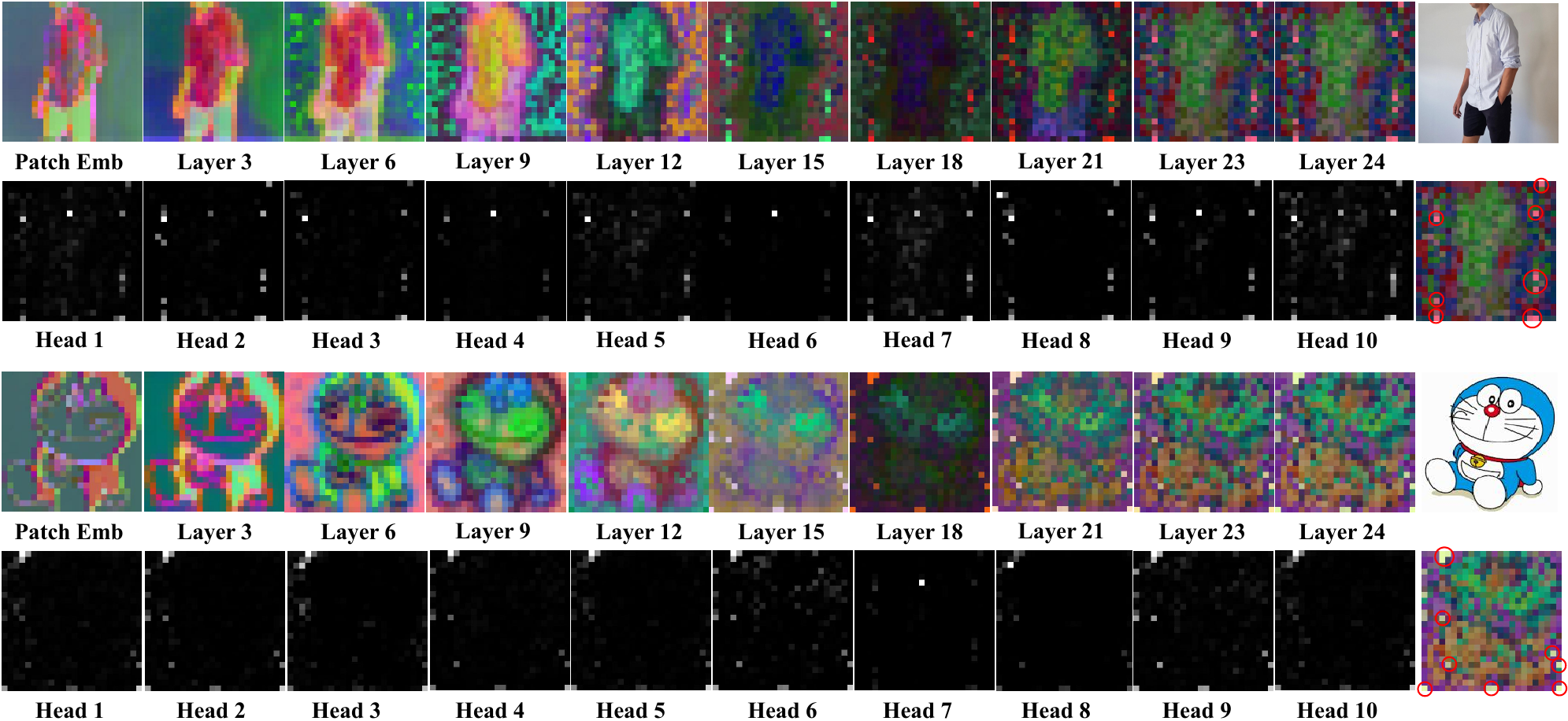}
  \caption{Visualizations of the visual feature map and attention map pertaining to the [CLS] token. Here we select 10 layers in Vision Transformer to show their output. We present the attention maps corresponding to the [CLS] token in the final layer. Notably, special tokens within both the feature map and attention map are identified using red circles. These marked points are referred to as ``visual anchors''. Details can be found in Section \ref{visual_anchor}.}
  \label{visual1}
\end{figure}
\label{visual_anchor}
Within the ViT, the input pixel-level features undergo a series of transformations. Understanding how the visual semantic is learned will bring us better insight of building the vision-language connector. To analyse in a more intuitive way, we visualize the feature map and attention map.

Given a set of Vision Transformer's feature maps, denoted as $\mathbf{V}\in \mathbb{R}^{N\times D}$, where $N$ represents the number of tokens and $D$ signifies the dimension of these tokens, we leverage dimension reduction for visual feature visualization. 
Initially, we extract all hidden states from the Vision Transformer. Subsequently, employing Principal Component Analysis (PCA) on each individual feature map, we derive low-dimensional features. We select the first three principal components to yield $\mathbf{V}'\in \mathbb{R}^{N\times 3}$, followed by normalizing the pixel values to the range of $[0, 255]$. 
To visually represent these features, we construct an image of the input size. Each patch within the image is then encoded using the obtained three-dimensional representation, effectively encapsulating the corresponding region's value within the image. By this way, we obtain the visualization of the features, as depicted in Figure \ref{visual1} (Row 1 and Row 3). Regarding the attention map, we derive it by extracting the attention weights associated with the last layer's [CLS] token. These attention scores reflect the significance of respective tokens. We exclude the attention directed towards the [CLS] token itself, yielding the attention map $\mathbf{A}\in \mathbb{R}^{H\times N}$, where $H$ and $N$ denote the number of attention heads and visual tokens. This process is depicted in Figure \ref{visual1}(Row 2 and Row 4).

Upon visual inspection, several noteworthy observations come to light. Specifically, the transformation of visual information exhibits a gradual obscuration of the feature map, accompanied by an increasing activation of specific tokens (depicted as pink and light yellow patches in the first and third lines, respectively). Initially, these activated tokens are just parts of the background, appearing indistinguishable from surrounding elements. However, over time, they progressively differentiate themselves from the background. Nevertheless, this evolutionary process lacks a discernible pattern. Examining the attention map, the [CLS] token, commonly used for aggregate global information of the input image, is anticipated to attend to the most salient regions. Paradoxically, the attention map of the [CLS] token predominantly focuses on a limited subset of tokens. We calculate the overlap among the activated tokens in feature map and the salient regions in attention map. The ratio reaches up to $\mathbf{69.38\%}$ (500 images are sampled for calculation), conjecturing that this alignment is not coincidental. For further validation, we visualize the pre-trained MLLMs text generation attention matrix (text to visual tokens), result can be found in Figure \ref{flamingo_attn}.

Based on above observation, we name these tokens ``visual anchors'' and assume them serving as pivotal points for information aggregation during the transformation of visual features. While the [CLS] token indeed integrates visual information, its reliance alone proves insufficient, necessitating the involvement of other visual anchors in conveying information to the [CLS] token. Consequently, this process facilitates the extraction of meaningful representations from the image.

\begin{figure}[h]
  \centering
  \includegraphics[width=14cm]{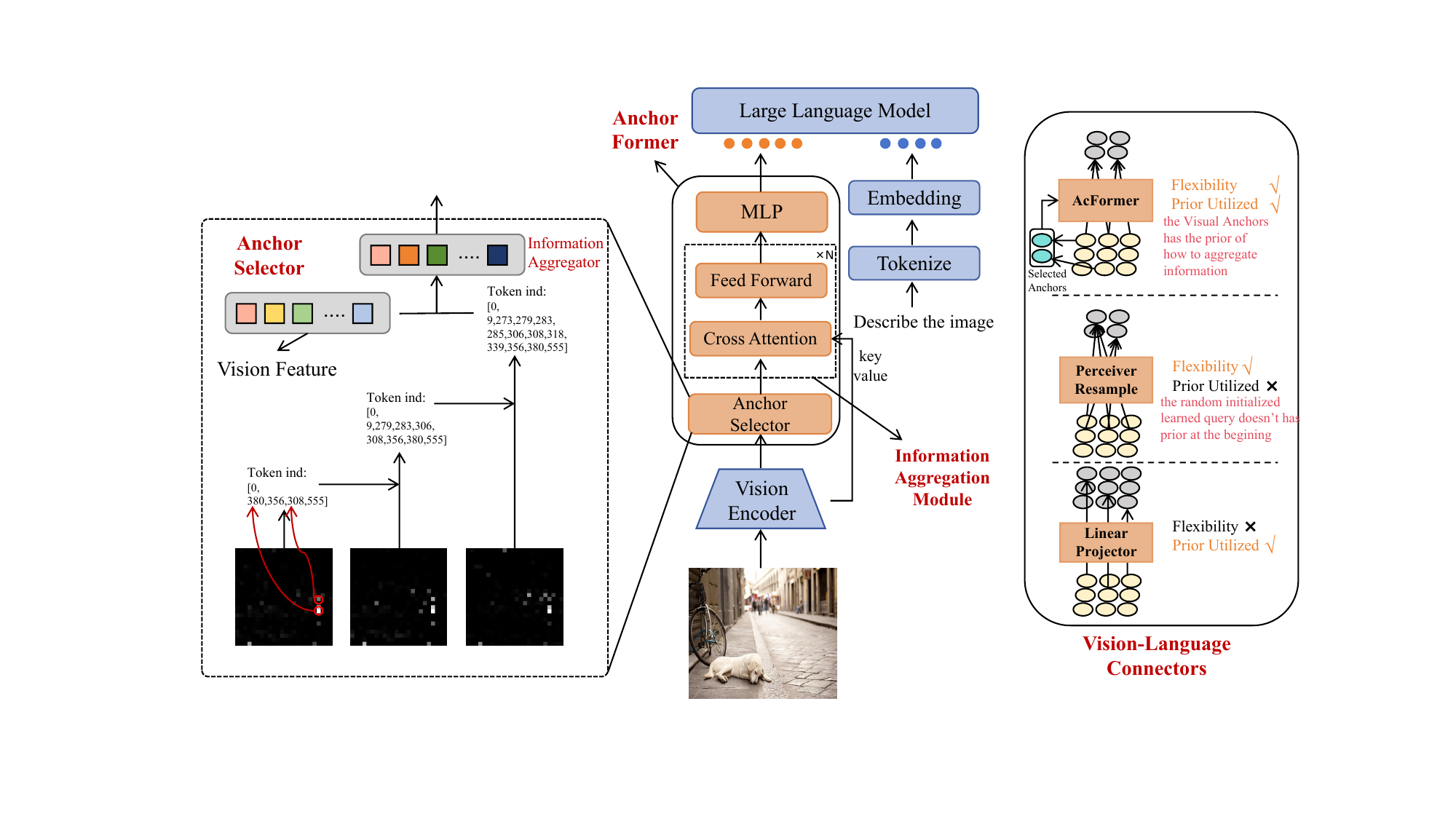}
  \caption{Visualization of Anchor Former (AcFormer). We propose our token selection algorithm code in detail at Section \ref{detailed_alg}.}
  \label{method}
\end{figure}

\subsection{Anchor Former}
As illustrated above, the vision information is aggregated through visual anchors. Similarly, within Multimodal Large Language Models (MLLMs), structures such as Q-Former and Perceiver Resampler also leverage information aggregation modules. However, these models use learnable queries as Information Aggregator to extract information from the visual feature map produced by a pre-trained Vision Transformer. In this method, the same queries are used for all images, which can lead to two issues. First, their effectiveness requires extensive datasets for training as demonstrated in Table \ref{ablation_scale}. Second, they may lead to  significant information loss, as illustrated in \cite{cha2023honeybee, ren2024grounding, huang2024dq}.

To address the aforementioned issues, we propose Anchor Former (AcFormer), which consists of Anchor Selector, Information Aggregation Module and Linear Projector. We show a rough view of Anchor Former in Figure \ref{method}. Our approach integrates insights from visual anchors with the established framework of the Perceiver Resampler. A key innovation in our method is the use of visual anchors associated with each image as Information Aggregator for aggregating visual information. The effectiveness of our approach relies on the selection of these salient points. This visual anchors allows for more precise and specific information extraction, improving both the accuracy and efficiency of the model.

\paragraph{Anchor Selector.} To avoid additional computation, we leverage the attention map of the [CLS] token for visual anchor selection. We introduce a progressive search algorithm to construct the Anchor Selector. Assuming the attention map is $\mathbf{A}\in \mathbb{R}^{H\times (N-1)}$, where $H$ represents the number of attention heads and $N-1$ denotes the number of visual tokens ([CLS] excluded). Let the token index list be $TL$, initially containing only the index 0, corresponding to the [CLS] token, an essential anchor. Suppose we still need $T_N$ tokens in addition to the [CLS] token. We select these tokens head by head, assuming each head will provide $\frac{T_N}{H}$ tokens. For each head, , we first sort the indices of the visual tokens based on their attention scores. Then, we select the top $\frac{T_N}{H}$ tokens from this sorted list. If the chosen token is already in $TL$, we choose the next token in the sorted order until we have the required number of unique tokens. We provide detailed algorithm in Figure \ref{method_detail}.

\paragraph{Information Aggregation Module.} In our approach, we employ selected visual anchors as Information Aggregator, combined with a cross-attention module to aggregate information. Let the Information Aggregator be denoted as $\mathbf{IA}\in \mathbb{R}^{(T_N+1)\times D}$ and the origin visual tokens as $\mathbf{R}_v\in \mathbb{R}^{N\times D}$. Our proposed model, Information Aggregation Module, is a bidirectional transformer encoder. We denote the cross-attention module as Attn and the feedforward module as FF. Let LN represent layer normalization. For a single layer in the model, the operations are as follows:
\begin{align}
    \mathbf{A}_{out} = \mathbf{IA} + \text{Attn}(\text{query=}\text{LN}(\mathbf{IA}), \text{key=}\text{LN}(\mathbf{R}_{v}),\text{value=}\text{LN}(\mathbf{R}_{v})),
\end{align}
\begin{align}
    \mathbf{H} = \mathbf{A}_{out} + \text{FF}(\text{LN}(\mathbf{A}_{out})),
\end{align}
Where $\mathbf{H}$ means the hidden states. We use $\mathbf{H}_v$ to represent the last hidden states. 
Let $\mathbf{T}$ represents text embedding. We use the Proj to represent the Multilayer Perception. We obtain the final multimodal input embeddings for LLM as bellow,
\begin{align}
    \mathbf{LM}_{in} = \text{Concat}(\text{Proj}(\mathbf{H}_v), \mathbf{T}).
\end{align}
\section{Experiments}
\subsection{Settings}
\paragraph{Benchmarks.} We employ nine distinct benchmarks to comprehensively assess the overall efficacy of our proposed method. The specifics of these benchmarks are delineated in Table \ref{bench_detail}. Notably, in our experimental setup, as we reduce the number of image tokens, our focus is primarily directed towards enhancing visual perception capabilities. 
As a result, we pay particular attention to benchmarks such as TextVQA, which challenged the model's fine-grained visual perception ability \cite{hong2023cogagent, li2023monkey}. 

\paragraph{\label{im_details1}Implementation details.} In our experimental setup, we utilize 7B and 13B Vicuna-v1.5 as Large Language Models (LLMs) \cite{vicuna2023}. The CLIP ViT-L/14 model, pre-trained with a resolution of 336, serves as our vision encoder. We select the last but one layer's output from the vision encoder as our vision feature. For Anchor Former, we configure it with 6 layers and a hidden dimension of 512, employing 8 attention heads, each with a dimension of 64. The feedforward module utilize 2048 as the hidden dimension. Regarding the training dataset, we leverage the dataset utilized in LLaVA-1.5 \cite{liu2023improvedllava}, with 558k samples for pre-training and 665k samples for instruction tuning. Our experimentation encompasses the evaluation of various vision-language connectors, including the Perceiver Resampler, Anchor Former, pooling, C-Abstractor and utilizing pooled tokens as queries for the Perceiver Resampler. To maintain consistency, we construct our model using the official code provided by LLaVA. Specifically, we adjust the pre-training initial learning rate from $1e^{-3}$ to $5e^{-4}$. Other configuration is the same with origin LLaVA-1.5.

\subsection{Main Results}
We present the main results in Tables \ref{main_result1} and \ref{main_result2}, organized by benchmark type. Upon observation of these tables, it becomes apparent that our model achieves robust performance despite being trained with limited data and visual tokens. Notably, even with only 145 or 257 tokens, our model achieves performance comparable to that of the original LLaVA-1.5 model, which utilizes 577 visual tokens as input. This performance holds across various benchmarks, including those that require high-level visual perception (e.g., VisWiz, TextVQA) and those that assess overall capability (e.g., MME, GQA). 

However, it should be noted that although our model is overall effective, it performs slightly worse than LLaVA-1.5 on certain benchmarks such as GQA and VQAv2. Given that our method only applies significantly fewer visual tokens (less than half), the slightly performance drop meets expectations. Nevertheless, the performance gap is relatively small and can be considered marginal in light of the substantial increase in speed.

\begin{table}
\centering
\caption{Results on benchmark designed for MLLMs. V-T Num means the visual tokens number. V-T Num influences the computation cost that the bigger the V-T Num the heavier the computation cost is. Speed here means the relative pre-training speed with respect to LLaVA-1.5.}
\label{main_result1}
\begin{adjustbox}{width=1.0\textwidth}
\begin{tabular}{l|c|c|c|c|cccc|c}
\toprule
Model  & LLM &Connector &V-T Num &Res & POPE& MME& MMB& MM-Vet& Speed ($\uparrow$)\\
\midrule
\rowcolor{gray!25}
\multicolumn{10}{l}{Approaches using 7B Large Language Models}\\
\midrule
MiniGPT-4  \cite{zhu2023minigpt4}         & Vicuna-7B & Resampler    & 32  & 224& 72.2 & 726.0 & 24.3 & 22.1& -\\
mPLUG-Owl2\cite{ye2024mplugowl}           & LLaMA2-7B & Resampler    & 32  & 224& -    & 1243.4& 49.4 & -   & -\\
InstructBLIP\cite{dai2023instructblip}    & LLaMA2-7B & Q-Former     & 32  & 224& 78.9 & -     & 36.0 & 26.2& -\\
\midrule
LLaVA (v1) \cite{liu2024visual}           & LLaMA-7B  & Linear       & 257 & 224& 67.7 & 717.5 & 38.7 & -   & -\\
LLaMA-AdapterV2 \cite{gao2023llamaadapter}& LLaMA2-7B & LLaMA-Adapter& 257 & 224& -    & 1221.6& 41.0 & 31.4& -\\
Shikra \cite{chen2023shikra}              & Vicuna-7B & Linear       & 257 & 224& -    & -     & 58.8 & -   & -\\
Qwen-VL\cite{bai2023qwenvl}               & Qwen-7B   & Resampler    & 256 & 448& -    & -     & 38.2 & -   & -\\
Qwen-VL-Chat\cite{bai2023qwenvl}          & Qwen-7B   & Resampler    & 256 & 448& -    & 1845.3& 60.6 & -   & -\\
\midrule
LLaVA-1.5 \cite{liu2023improvedllava}     & Vicuna-7B & Linear       & 577 & 336& 85.9 & 1794.6& 64.3 & \textbf{30.5}& $1.00\times$\\
\midrule
Ours& Vicuna-7B & AcFormer       & 145 & 336& \textbf{86.4} & \textbf{1846.1}& \textbf{68.4} & 30.3& $\mathbf{2.23\times}$\\
\midrule
\rowcolor{gray!25}
\multicolumn{10}{l}{Approaches using 13B Large Language Models}\\
\midrule
MiniGPT-4  \cite{zhu2023minigpt4}         & Vicuna-13B & Resampler    & 32  & 224& -    & 1158.7& -    & 24.4& -\\
InstructBLIP\cite{dai2023instructblip}    & Vicuna-13B & Q-Former     & 32  & 224& 78.9 & 1504.6& -    & 25.6& -\\
BLIP-2\cite{li2023blip}                   & Vicuna-13B & Q-Former     & 32  & 224& 85.3 & -     & -    & 22.4& -\\
\midrule
LLaVA-1.5 \cite{liu2023improvedllava}     & Vicuna-13B & Linear       & 577 & 336& 85.9 & 1826.7& 67.7 & \textbf{35.4}& $1.00\times$\\
\midrule
Ours& Vicuna-13B & AcFormer       & 145 & 336& \textbf{86.1} & \textbf{1870.0}& \textbf{69.2 }& 34.1& $\mathbf{2.30\times}$\\
\bottomrule
\end{tabular}
\end{adjustbox}
\end{table}

\begin{table}[h]
\centering
\caption{Results on General VQA tasks. V-T Num means the visual tokens number. V-T Num influences the computation cost that the bigger the V-T Num the heavier the computation cost is. Speed here means the relative pre-training speed with respect to LLaVA-1.5.}
\label{main_result2}
\begin{adjustbox}{width=1.0\textwidth}
\begin{tabular}{l|c|c|c|c|ccccc|c}
\toprule
Model  & LLM &Connector &V-T Num &Res & TextVQA& GQA& VQAv2& VisWiz& SQA$_{img}$ & Speed ($\uparrow$)\\
\midrule
\rowcolor{gray!25}
\multicolumn{11}{l}{Approaches using 7B Large Language Models}\\
\midrule
InstructBLIP\cite{dai2023instructblip}    & LLaMA2-7B & Q-Former     & 32  & 224& -    & 49.2 & -    & 34.5 & 60.5 & -\\
\midrule
Shikra \cite{chen2023shikra}              & Vicuna-7B & Linear       & 257 & 224& -    & -    & 77.4 & -    & -  & -\\
IDEFICS-9B \cite{laurenccon2024obelics}   & LLaMA-7B  & Cross Attn   & 257 & 224& -    & 38.4 & 50.9 & 35.5 & -  & -\\
Qwen-VL\cite{bai2023qwenvl}               & Qwen-7B   & Resampler    & 256 & 448& -    & 59.3 & \textbf{78.8} & 35.2 & 67.1 & -\\
Qwen-VL-Chat\cite{bai2023qwenvl}          & Qwen-7B   & Resampler    & 256 & 448& -    & 57.5 & 78.2 & 38.9 & 68.2 & -\\
\midrule
LLaVA-1.5 \cite{liu2023improvedllava}     & Vicuna-7B & Linear       & 577 & 336& 58.2 & \textbf{62.0} & 78.5 & 50.0 & 66.8 & $1.00\times$\\
\midrule
Ours& Vicuna-7B & AcFormer       & 257 & 336& \textbf{58.2} & 61.2& 78.4 & \textbf{52.8} & \textbf{69.4} & $\mathbf{1.65\times}$\\
\midrule
\rowcolor{gray!25}
\multicolumn{11}{l}{Approaches using 13B Large Language Models}\\
\midrule
InstructBLIP\cite{dai2023instructblip}    & Vicuna-13B & Q-Former     & 32  & 224& -    & 49.5  & -    & 33.4 & 63.1 & -\\
BLIP-2\cite{li2023blip}                   & Vicuna-13B & Q-Former     & 32  & 224& -    & 41.0  & 41.0 & 19.5 & 61.0 & -\\
\midrule
LLaVA-1.5 \cite{liu2023improvedllava}     & Vicuna-13B & Linear       & 577 & 336& 61.2 & \textbf{63.3}  & \textbf{80.0} & 53.6 & 71.6 & $1.00\times$\\
\midrule
Ours& Vicuna-13B & AcFormer       & 257 & 336& \textbf{61.3} & 63.0 & 79.8 & \textbf{53.7} & \textbf{71.8} & $\mathbf{1.69\times}$\\
\bottomrule
\end{tabular}
\end{adjustbox}
\end{table}

\subsection{Ablation Results}
\begin{table}
\centering
\caption{Ablation studies. ``Pooling'' denotes direct pooling of visual token. ``Pooling-PR'' employs the pooled tokens as queries for the Perceiver Resampler. ``Random-PR'' means the Perceiver Resampler using randomly selected tokens from the vision feature map as query. ``PR'' refers to the Perceiver Resampler using learnable queries. ``AcFormer'' represents our proposed Anchor Former. The configuration of the C-Abstractor follows Honeybee \cite{cha2023honeybee}. V-T Num means the visual tokens number.}
\label{ablation_res}
\begin{adjustbox}{width=0.85\textwidth}
\begin{tabular}{l|c|c|c|cccc}
\toprule
Model  & LLM &Connector &V-T Num. & TextVQA& GQA& MMB& MME\\
\midrule
\multirow{5}{*}{\centering LLaVA-1.5}& Vicuna-7B & Pooling            & 65  & 53.4 & 59.8& 66.8 & 1734.0\\
& Vicuna-7B & Pooling-PR         & 65  & 53.9 & \textbf{60.0}& 66.8 & 1728.9\\
& Vicuna-7B & Random-PR          & 65  & 53.9 & 59.1& 66.9 & 1728.7\\
& Vicuna-7B & PR          & 65  & 51.0 & 56.1& 63.2 & 1702.8\\
& Vicuna-7B & C-Abstractor       & 65  & 52.8 & 59.0& 67.0 & 1743.3\\
& Vicuna-7B & AcFormer       & 65  & \textbf{56.1} & 59.2& \textbf{67.3} & \textbf{1744.2}\\
\midrule
\multirow{6}{*}{\centering LLaVA-1.5}& Vicuna-7B & Pooling            & 145 & 55.1 & 60.9& 68.0 & 1791.4\\
& Vicuna-7B & Pooling-PR         & 145 & 54.7 & 60.9& 68.0 & 1759.1\\
& Vicuna-7B & Random-PR          & 145 & 54.6 & 59.7& 67.0 & 1772.7\\
& Vicuna-7B & PR          & 145 & 52.1 & 56.4& 65.4 & 1720.8\\
& Vicuna-7B & C-Abstractor       & 145 & 53.4 & 60.2& 67.8 & 1775.4\\
& Vicuna-7B & AcFormer       & 145 & \textbf{58.0} & \textbf{61.3}& \textbf{68.4} & \textbf{1846.1}\\
\midrule
\multirow{3}{*}{\centering LLaVA-1.5}& Vicuna-7B & PR           & 257 & 52.3 & 56.8& 65.7 & 1735.9\\
& Vicuna-7B & C-Abstractor        & 257 & 53.7 & 60.8& 68.3 & 1790.0\\
& Vicuna-7B & AcFormer        & 257 & \textbf{58.2} & \textbf{61.2}& \textbf{68.3} & \textbf{1848.8}\\
\midrule
\multirow{3}{*}{\centering LLaVA-1.5}& Vicuna-13B & PR          & 145 & 53.4 & 56.9& 64.7 & 1749.3\\
& Vicuna-13B & C-Abstractor       & 145 & 58.5 & 62.1& 68.8 & 1823.6\\
& Vicuna-13B & AcFormer       & 145 & \textbf{60.7} & \textbf{62.8}& \textbf{69.2} & \textbf{1869.3}\\
\bottomrule
\end{tabular}
\end{adjustbox}
\end{table}

\begin{table}
\centering
\caption{Ablation studies on whether to directly use the selected tokens as input.}
\label{ablation_res_pr}
\begin{adjustbox}{width=0.85\textwidth}
\begin{tabular}{l|c|c|c|cccc}
\toprule
Model  & LLM &Connector &V-T Num. & TextVQA& GQA& MMB& MME\\
\midrule
\multirow{3}{*}{\centering LLaVA-1.5}& Vicuna-7B & Top-P        & 145  & 56.3 & 60.8& 68.2 & 1798.8\\
& Vicuna-7B & E-ViT        & 146  & 57.1 & 61.0& 68.3 & 1808.4\\
& Vicuna-7B & AcFormer      & 145  & \textbf{58.0} & \textbf{61.3}& \textbf{68.4} & \textbf{1846.1}\\
\bottomrule
\end{tabular}
\end{adjustbox}
\end{table}
We mainly compare different visual Connectors. To facilitate understanding, we provide definitions for some terms. Pooling denotes direct pooling of visual token. Pooling-PR employs the pooled tokens as queries for the Perceiver Resampler. Random-PR means the Perceiver Resampler using randomly selected tokens from the vision feature map as query. PR refers to the commonly used Perceiver Resampler. Our experimental findings validate the efficacy of our model from multiple perspectives. A comparison between PR and AcFormer indicates that the observed enhancement does not stem solely from an increase in \textbf{trainable parameters}. Additionally, comparison between AcFormer and Random-PR underscores the critical role of \textbf{Anchor Selector} in model performance. Furthermore, our evaluation of AcFormer against C-Abstractor reveals that the significance of inductive bias diminishes, as spatial coherence can be effectively maintained through cross-attention mechanisms within the Perceiver Resampler.

\paragraph{C-Abstractor.} 
We trained the model with training data from LLaVA-1.5. Our findings, as presented in Table \ref{ablation_res}, indicate that the C-Abstractor method achieves comparable performance to AcFormer across most benchmarks. However, in tasks such as TextVQA, which necessitates high-level visual perception (often demanding fine-grained visual analysis), C-Abstractor exhibits inferior performance. This empirical evidence underscores the efficacy of our proposed AcFormer. 

\paragraph{Pooling.} One immediate consideration is to aggregate visual tokens based on their spatial positions by pooling and combine them with the [CLS] token to form the input visual features. Our empirical investigation reveals that direct pooling emerges as a potent technique for compressing visual information. However, it fails in TextVQA, leading to around 3 points drop. 

\paragraph{Pooling-PR.} Given our direct utilization of pooled tokens within Large Language Models, a concern arises regarding whether the observed degradation stems from the reduction in trainable parameters within the Anchor Former. To address this concern, we conducted additional experiments wherein the pooled tokens served as queries for the Perceiver Resampler, termed as Pooling-PR. Examination of the results in the table reveals that this approach yielded even poorer performance compared to directly inputting the pooled tokens.

\paragraph{Random-PR.} While our method demonstrates superior performance compared with other approaches, a lingering question pertains to whether this improvement indeed stems from the Anchor Former. To address the concern, we randomly select tokens from the vision feature map  as the Information Aggregator. The empirical results presented in Table \ref{ablation_res} corroborate the efficacy of our proposed anchor selection method, substantiating its discernible impact on model performance. 

\paragraph{PR.} While substantiated the effectiveness of AcFormer, an unresolved question pertains to its superiority over the commonly used Perceiver Resampler, which employs randomly initiallized learnable queries for visual information aggregation. Examination of the table reveals that the Perceiver Resampler exhibits the poorest performance among the aforementioned methods. This observation underscores the diminished performance of the structure under conditions of limited training data.

\paragraph{Anchor Direct-in.} A related study \cite{haurum2023tokens} employ Top-K selection methodology for token compression. This approach shares similar token selection method with ours and involves the direct utilization of selected tokens as representations (without further cross attention) for visual information. However, it primarily caters to classification tasks, which prioritize global understanding, potentially rendering it less suitable for tasks requiring nuanced comprehension. Noteworthy is the proposition by the authors of Haurum \cite{haurum2023tokens} regarding EViT. They employ the selected tokens and incorporate pooling mechanisms for the remaining tokens as input. The findings, as detailed in Table \ref{ablation_res_pr}, empirically underscore the significance of retaining unselected tokens, as they still encapsulate valuable information.

\begin{table}[h]
\centering
\caption{Ablation studies on the visual connector when scaling up the training data.}.
\label{ablation_scale}
\begin{adjustbox}{width=1.0\textwidth}
\begin{tabular}{l|c|c|c|cccccc}
\toprule
Model  & LLM &Connector &V-T Num. & TextVQA& GQA& OKVQA& VQAv$_{2}$&VizWiz&MME\\
\midrule
\rowcolor{gray!25}
\multicolumn{10}{l}{\textbf{Pretrain Dataset}: 60M image-text pairs from LAION-115M, COYO, LAION COCO}\\
\rowcolor{gray!25}
\multicolumn{10}{l}{\textbf{Instruction Finetuning Dataset}:LLaVA-665k}\\
\midrule
\multirow{2}{*}{\centering LLaVA-1.5}& OpenLLaMA-3B & PR        & 145  &35.03&54.35& 54.14 &70.04 & 31.16 & 1592.7\\
& OpenLLaMA 3B & AcFormer       & 145  & \textbf{35.89} & \textbf{55.45}& \textbf{55.15} & \textbf{72.76} & \textbf{33.88} & \textbf{1622.3}\\
\midrule
\rowcolor{gray!25}
\multicolumn{10}{l}{\textbf{Pretrain Dataset}: 60M image-text pairs from LAION-115M, COYO, LAION COCO}\\
\rowcolor{gray!25}
\multicolumn{10}{l}{\textbf{Instruction Finetuning Dataset}:Cauldron (roughly 1.8M)}\\
\midrule
\multirow{2}{*}{\centering LLaVA-1.5}& OpenLLaMA-3B & PR        & 145  &38.76&44.98& 49.66 &70.83 & 32.87 & 1502.5\\
& OpenLLaMA 3B & AcFormer       & 145  & \textbf{40.49} & \textbf{45.67}& \textbf{50.57} & \textbf{73.01} & \textbf{32.97} & \textbf{1523.1}\\
\midrule
\bottomrule
\end{tabular}
\end{adjustbox}
\end{table}
\subsection{Scaling Up the Training Dataset}
We conducted a data scaling experiment to further evaluating our proposed method's performance on large-scale data scene. Considering the expensive training cost, we replaced the Large Language Model with OpenLLaMA-3B \cite{openlm2023openllama}. Approximately 60 million image-text pairs were employed for pre-training, sampled from Laion115M, coyo-238M, and laion-coco100M datasets. For instruction fine-tuning, we utilized two datasets, covering LLaVA-665k and Cauldron \cite{laurençon2024matters}. Notably, Cauldron (1.8M) is a larger instruction finetuning dataset than LLaVA-665k (0.66M) for MLLMs.

Our focus was on comparing Perceiver Resampler with our proposed AcFormer. The results, summarized in Table \ref{ablation_scale}, consistently demonstrate the superiority of AcFormer over the Perceiver Resampler. This  illustrate that our method not only works well for the scene of limited data but also for larger data (both pre-train dataset and instruction tuning dataset).

\section{Conclusions}
In this study, we reveal the existence of the visual anchors and present the hypothesis that visual anchors are strong visual information aggregators. With the observation, we propose the Anchor Former, an effective vision-language connector. Notably, Anchor Former distinguishes itself from the conventional information aggregation methods (e.g., Q-Former and Perceiver Resampler) by adopting visual anchors as Information Aggregator. To build Anchor Former, we present the progressive search algorithm to effectively extract the visual anchors. Extensive experiments on different benchmarks consistently underscores the efficacy of Anchor Former in improving the models' accuracy while simultaneously removing the redundant visual tokens in MLLMs. 
\section*{Acknowledgements} 
This research is partially funded by National Natural Science Foundation of China (Grant No U21B2045, U20A20223), Beijing Nova Program (20230484276), and Youth Innovation Promotion Association CAS (Grant No. 2022132).

\clearpage

{\small
\bibliographystyle{ieee_fullname}
\bibliography{egbib}
}

\newpage
\appendix
\section{Visualization of The Feature From OpenAI CLIP}
\begin{figure}[h]
  \centering
  \includegraphics[width=14cm]{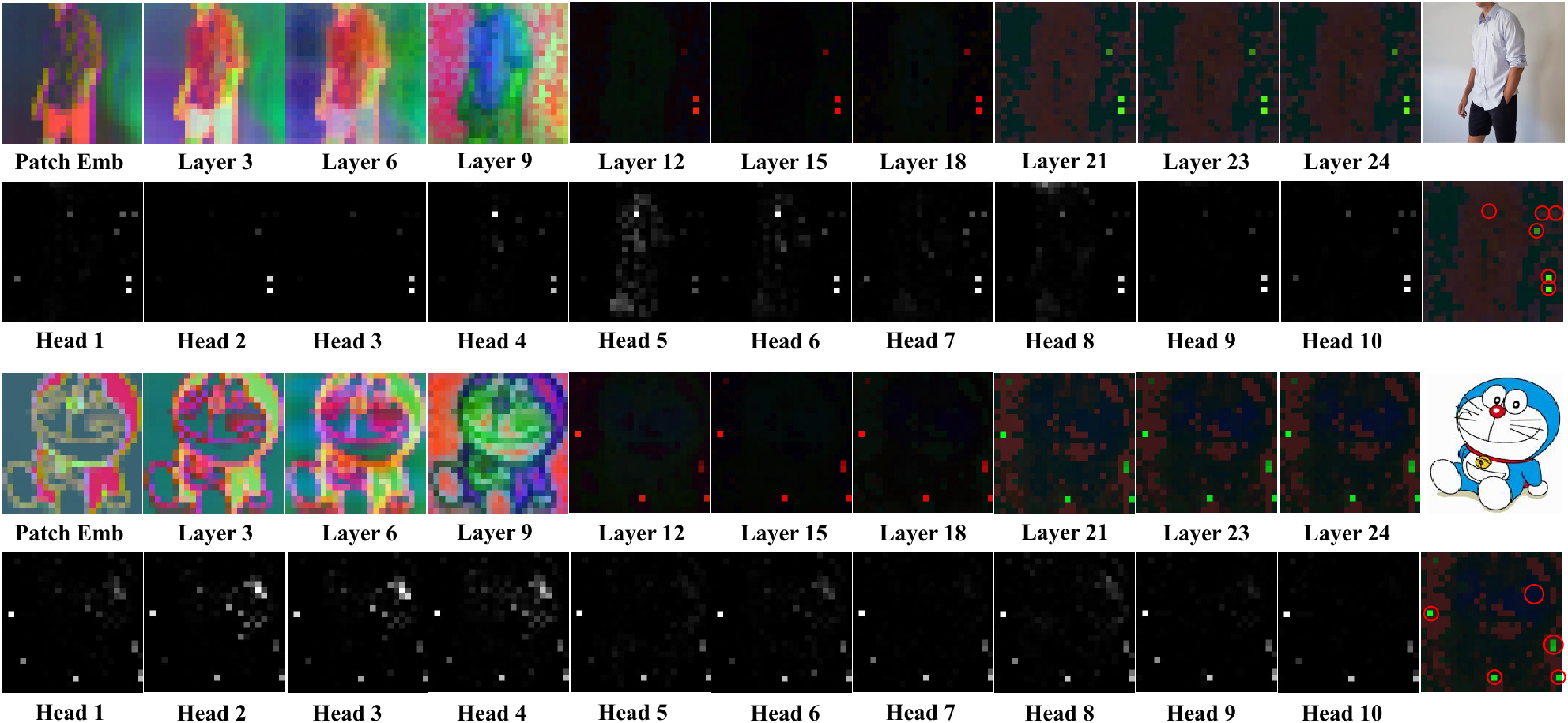}
  \caption{Visualization of the Openai CLIP. Figure \ref{visual1} shows the visualization of EVA CLIP. We show that the phenomenon happens in different CLIP.}
  \label{visual_openclip}
\end{figure}

We have shown the visualization of the feature map and attention map of EVA-CLIP-L in the main text. To validate whether other models (OpenAI-CLIP) also has the same characteristic, we offer analogous visualizations in Figure \ref{visual_openclip}, revealing comparable phenomena. While diverging from EVA-CLIP-L in certain aspects, OpenAI-CLIP-L nevertheless manifests certain anchor points. These shows that our observation is not an isolated case.

\section{Visualization of The Attention Map From The Pre-trained MLLMs}
To further substantiate our hypothesis, we conducted visualizations of the attention matrix of the pre-trained Multimodal Language Models (MLLMs). We believe directly obtaining the generated text's attention will give us better understanding of the visual anchors. Recognizing the potential generation of multiple tokens simultaneously, we adopted a methodology akin to LLaVA's approach, constraining the model to provide single-word or phrase responses for visualization of the attention mechanism on the answer (key word). Given the multilayered architecture of MLLMs, we specifically chose middle layers for visualization, as prior work \cite{chen2024image} has indicated their pivotal role in comprehending visual signals.

\begin{figure}
  \centering
  \includegraphics[width=14cm]{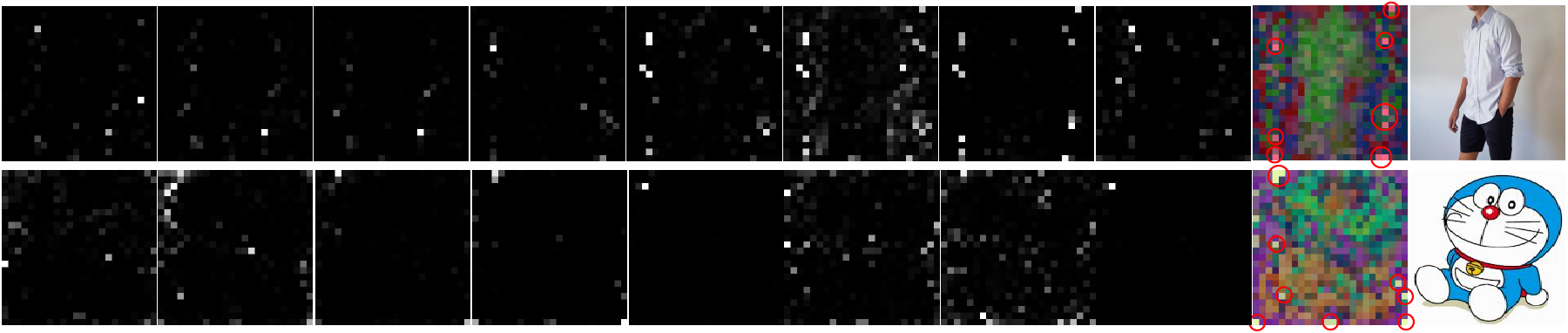}
  \caption{Visualization of the attention map from the pre-trained Flamingo model (Removing the Perceiver Resampler). The attention map is the generated text's attending to the corresponding image patches.}
  \label{flamingo_attn}
\end{figure}

Two types of MLLMs are considered, exemplified by Flamingo and LLaVA. As Flamingo segregates attention between text-only and image-text modalities, we opted for this model for visualization purposes. It is noteworthy that we excluded the Perceiver Resampler and directly leveraged all visual tokens to discern their relative importance. The visualization is accessible in Figure \ref{flamingo_attn}.

\section{More Samples For Visual Anchor Visualization}
\begin{figure}
  \centering
  \includegraphics[width=14cm]{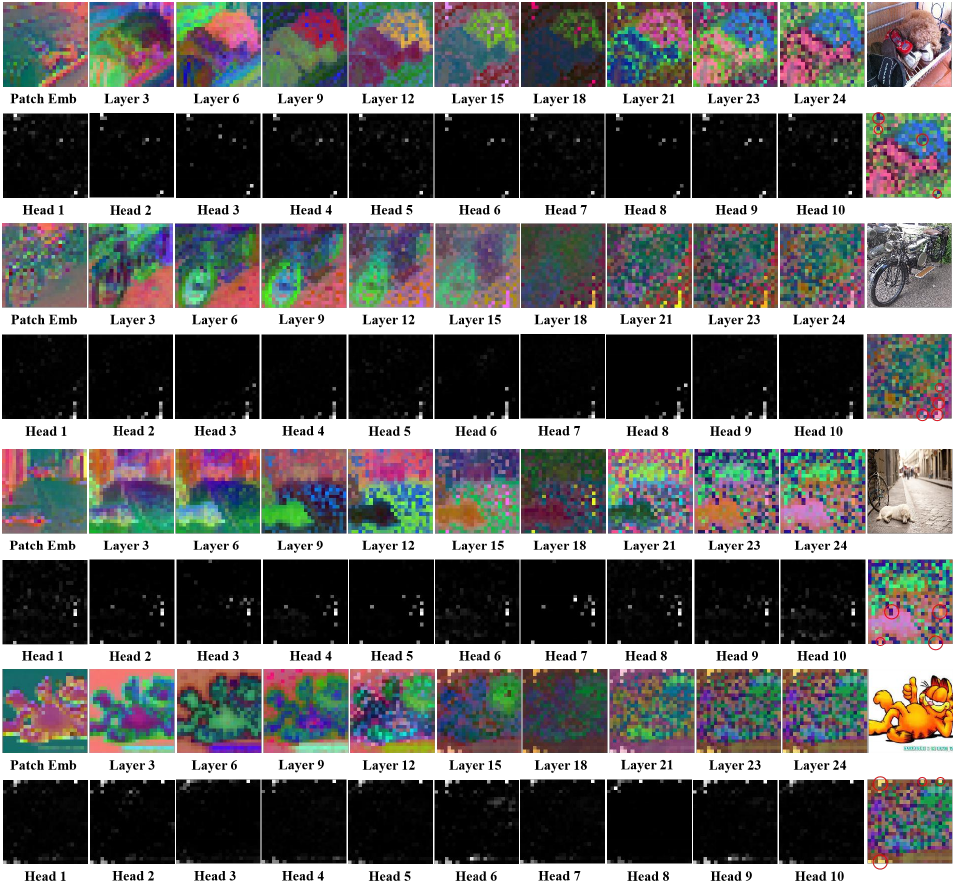}
  \caption{More samples for visualization of the visual anchors.} 
  \label{visual_more}
\end{figure}

We provide more visualization samples in Figure \ref{visual_more} to demonstrate that the phenomenon of vision anchors exists in both complex real-world scenes and animated scenes.

\section{Detailed Algorithm}
\label{detailed_alg}
We have provided a rough description in Figure \ref{method}. Here, we present the detailed Python code for token selection in Figure \ref{method_detail}. The complete code is available in the supplementary material. We use a top-k search to select visual tokens. Specifically, assume the attention matrix is $\mathbf{A} \in \mathbb{R}^{H \times 1 \times N}$, where $H$ is the number of heads and $N$ is the number of image patches. We traverse each head of the attention matrix, and for each head, we select $\frac{\mathbf{T-1}}{\mathbf{H}}$ tokens based on the sorted indices and already selected sequence to avoid duplication. Here, $\mathbf{T}$ represents the total number of desired tokens.
\begin{figure}
  \centering
  \includegraphics[width=13.5cm]{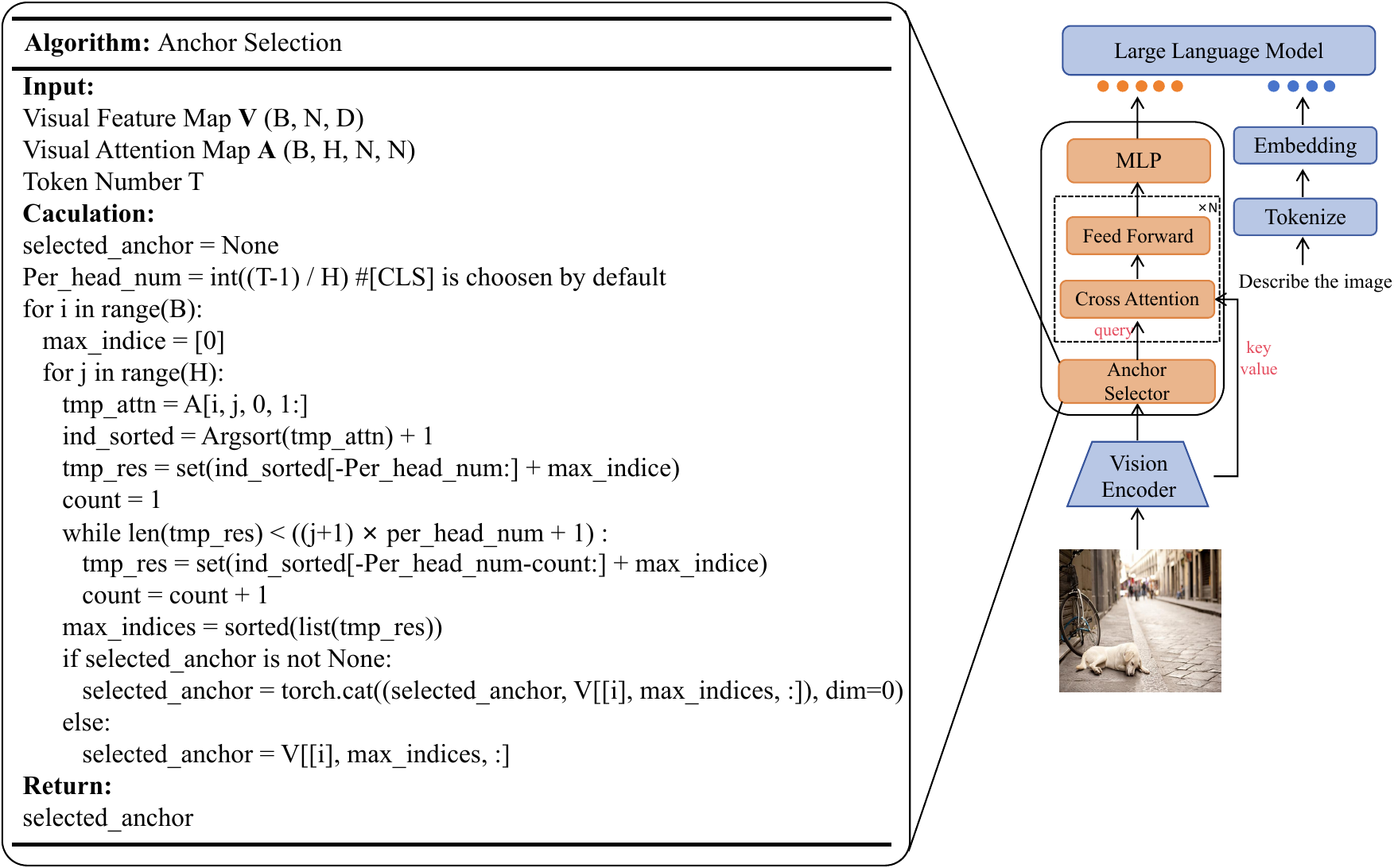}
  \caption{Detailed code for the anchor selector within Anchor Former (AcFormer).}
  \label{method_detail}
\end{figure}

\section{Training Resources}
We provide details for our training resource for main experiment in Table \ref{resource_used}. Notably, for the data scaling experiment, we utilize 16 Nvidia A100 (80G). It takes roughly 28 hours for pre-train and 5 hours for IFT. 

\begin{table}
\centering
\caption{Details on the training time. Pt Bz means the pre-train batch size. And IFT Bz means the instruction finetune batch size.}
\label{resource_used}
\begin{adjustbox}{width=1.0\textwidth}
\begin{tabular}{l|c|c|c|cc|cc}
\toprule
Methods &LLM & Training resource & Visual Token Num &Pt Bz &IFT Bz & Pt time & IFT time\\
\midrule
Pooling     & Vicuna-7B & 8 A100(80G)& 65& 256& 128& 39m& 9h25m\\
Pooling-PR  & Vicuna-7B & 8 A100(80G)& 65& 256& 128& 42m& 9h51m\\
Random-PR   & Vicuna-7B & 8 A100(80G)& 65& 256& 128& 42m& 9h52m\\
Origin-PR   & Vicuna-7B & 8 A100(80G)& 65& 256& 128& 42m& 9h52m\\
C-Abstractor& Vicuna-7B & 8 A100(80G)& 65& 256& 128& 41m& 9h51m\\
AcFormer      & Vicuna-7B & 8 A100(80G)& 65& 256& 128& 42m& 9h52m\\
\midrule
Pooling     & Vicuna-7B & 8 A100(80G)& 145& 256& 128& 1h03m & 10h13m\\
Pooling-PR  & Vicuna-7B & 8 A100(80G)& 145& 256& 128& 1h20m & 10h51m\\
Random-PR   & Vicuna-7B & 8 A100(80G)& 145& 256& 128& 1h19m & 10h50m\\
Origin-PR   & Vicuna-7B & 8 A100(80G)& 145& 256& 128& 1h20m & 10h52m\\
C-Abstractor& Vicuna-7B & 8 A100(80G)& 145& 256& 128& 1h18m & 10h50m\\
AcFormer      & Vicuna-7B & 8 A100(80G)& 145& 256& 128& 1h20m & 10h12m\\
\midrule
Origin-PR   & Vicuna-7B & 8 A100(80G)& 257& 256& 128& 1h50m & 10h29m\\
C-Abstractor& Vicuna-7B & 8 A100(80G)& 257& 256& 128& 1h49m & 10h25m\\
AcFormer      & Vicuna-7B & 8 A100(80G)& 257& 256& 128& 1h50m & 10h30m\\
\midrule
Origin-PR   & Vicuna-13B & 8 A100(80G)& 145& 256& 128& 2h04m & 17h31m\\
C-Abstractor& Vicuna-13B & 8 A100(80G)& 145& 256& 128& 2h02m & 17h29m\\
AcFormer      & Vicuna-13B & 8 A100(80G)& 145& 256& 128& 2h04m & 17h32m\\
\bottomrule
\end{tabular}
\end{adjustbox}
\end{table}

\section{Detailed Description of The Benchmarks}
\begin{table}
\centering
\caption{Details on the chosen benchmark.}
\label{bench_detail}
\begin{tabular}{c|l|c}
\toprule
Benchmark      & Description of the task & Metric \\
\midrule
TextVQA \cite{singh2019towards}   & QAs about text in image (Visual Perception)           & VQA score ($\uparrow$)  \\
VizWiz VQA \cite{gurari2018vizwiz}& QAs about image from blinds (Visual Perception)       & VQA score ($\uparrow$)  \\
GQA \cite{hudson2019gqa}          & QAs of real world comprehension and complex reasoning & EM ($\uparrow$)         \\
VQAv2 \cite{VQA}                  & QAs require vision, language and prior world knowledge& VQA score ($\uparrow$)  \\
POPE  \cite{li2023evaluating}     & QAs for Object Hallucination evaluation               & F1 Score  ($\uparrow$)  \\
Sci-QA(Img) \cite{lu2022learn}    & QAs about Science                                     & Accuracy ($\uparrow$)   \\
MME \cite{fu2023mme}              & Comprehensive Evaluation Benchmark for MLLMs          & Accuracy ($\uparrow$)   \\
MMbench \cite{liu2023mmbench}     & Comprehensive Evaluation Benchmark for MLLMs          & Accuracy ($\uparrow$)   \\
MM-Vet \cite{yu2023mmvet}         & Integrated Capabilities Benchmark for MLLMs           & GPT-4 score($\uparrow$) \\
\bottomrule
\end{tabular}
\end{table}
There exist numerous benchmarks for assessing the proficiency of Multi-Modal Large Language Models (MLLMs), each imbued with its own inherent biases. For instance, benchmarks such as OK-VQA \cite{marino2019ok} primarily concentrate on appraising the model's pre-existing knowledge base. Conversely, benchmarks like TextVQA \cite{singh2019towards} and VizWiz-VQA \cite{gurari2018vizwiz} scrutinize the models' prowess in visual perception. Notably, a plethora of newly introduced benchmarks tailored specifically for MLLMs have surfaced. MME \cite{fu2024mme} and MMBench \cite{liu2023mmbench} stand out as comprehensive benchmarks aimed at gauging the overall performance of MLLMs. However, they mandate MLLMs to furnish responses in binary or multiple-choice formats. POPE \cite{li2023evaluating} has been devised to assess MLLMs' propensity for hallucination. MM-VET \cite{yu2023mm} and InfiMM \cite{han2023infimm} scrutinize the open-ended question-answering capability of MLLMs, aided by the assistance of GPT-4.

\section{Limitations}
\label{limitations}
Although we have conducted extensive experiments, there are still aspects requiring further investigation. For instance, the utilization of larger training datasets with corresponding larger models remains unexplored due to resource constraints in our experiments. Additionally, more theoretical analysis are needed for better elucidating the underlying reasons for the emergence of these visual anchors.

\section{Broader Impacts}
\label{broader impact}

Our model, despite its capabilities, may encounter certain risks. As it is built upon LLaMA, Vicuna, and CLIP, it inherits some issues associated with large language models (LLMs) and vision encoders. One significant risk is \textbf{hallucination}, where the model might generate content that contradicts the facts. This poses a concern, especially when applied in critical fields such as medicine. Additionally, \textbf{biases} present in the LLM and vision encoder (CLIP) could be transferred to our model, potentially resulting in biased outputs. Lastly, while \textbf{energy consumption} is not a primary concern due to the smaller pretraining dataset used, it may become an issue when scaling up the pretraining dataset or increasing the model size.
\newpage
\section*{NeurIPS Paper Checklist}

\begin{enumerate}

\item {\bf Claims}
    \item[] Question: Do the main claims made in the abstract and introduction accurately reflect the paper's contributions and scope?
    \item[] Answer: \answerYes{} 
    \item[] Justification: We give main claims accurately in abstract and instruction.

\item {\bf Limitations}
    \item[] Question: Does the paper discuss the limitations of the work performed by the authors?
    \item[] Answer: \answerYes{} 
    \item[] Justification: We discuss the limitations at section \ref{limitations}.

\item {\bf Theory Assumptions and Proofs}
    \item[] Question: For each theoretical result, does the paper provide the full set of assumptions and a complete (and correct) proof?
    \item[] Answer: \answerNo{} 
    \item[] Justification: This is an \textbf{observational and experimentally validated study}, with no theoretical argumentation involved.

    \item {\bf Experimental Result Reproducibility}
    \item[] Question: Does the paper fully disclose all the information needed to reproduce the main experimental results of the paper to the extent that it affects the main claims and/or conclusions of the paper (regardless of whether the code and data are provided or not)?
    \item[] Answer: \answerYes{} 
    \item[] Justification: This work can be fully reproducted with the information from the paper. We provide the main code in the appendix \ref{method_detail} in detail.

\item {\bf Open access to data and code}
    \item[] Question: Does the paper provide open access to the data and code, with sufficient instructions to faithfully reproduce the main experimental results, as described in supplemental material?
    \item[] Answer: \answerYes{} 
    \item[] Justification: Our model is built based on the open-sourced LLaVA, so our provided code in Figure \ref{method_detail} is enough for reproduction of the method.

\item {\bf Experimental Setting/Details}
    \item[] Question: Does the paper specify all the training and test details (e.g., data splits, hyperparameters, how they were chosen, type of optimizer, etc.) necessary to understand the results?
    \item[] Answer: \answerYes{} 
    \item[] Justification: We provide the implementation details in paragraph \ref{im_details1}.

\item {\bf Experiment Statistical Significance}
    \item[] Question: Does the paper report error bars suitably and correctly defined or other appropriate information about the statistical significance of the experiments?
    \item[] Answer: \answerNo{} 
    \item[] Justification: We experiment on many different benchmarks to eliminate statistical error.

\item {\bf Experiments Compute Resources}
    \item[] Question: For each experiment, does the paper provide sufficient information on the computer resources (type of compute workers, memory, time of execution) needed to reproduce the experiments?
    \item[] Answer: \answerYes{} 
    \item[] Justification: We report the training resources in section Table \ref{resource_used}.
    
\item {\bf Code Of Ethics}
    \item[] Question: Does the research conducted in the paper conform, in every respect, with the NeurIPS Code of Ethics \url{https://neurips.cc/public/EthicsGuidelines}?
    \item[] Answer: \answerYes{} 
    \item[] Justification: There is no bias or other special about the research.

\item {\bf Broader Impacts}
    \item[] Question: Does the paper discuss both potential positive societal impacts and negative societal impacts of the work performed?
    \item[] Answer: \answerYes{} 
    \item[] Justification: We provide Broader Impacts in section \ref{broader impact}.
    
\item {\bf Safeguards}
    \item[] Question: Does the paper describe safeguards that have been put in place for responsible release of data or models that have a high risk for misuse (e.g., pretrained language models, image generators, or scraped datasets)?
    \item[] Answer: \answerYes{} 
    \item[] Justification: Our proposed method is easy for reproduct that we will not release the model.

\item {\bf Licenses for existing assets}
    \item[] Question: Are the creators or original owners of assets (e.g., code, data, models), used in the paper, properly credited and are the license and terms of use explicitly mentioned and properly respected?
    \item[] Answer: \answerYes{} 
    \item[] Justification: We follow the license of the data and code used in our experiment.

\item {\bf New Assets}
    \item[] Question: Are new assets introduced in the paper well documented and is the documentation provided alongside the assets?
    \item[] Answer: \answerNA{} 
    \item[] Justification: There are not any asset in the paper.

\item {\bf Crowdsourcing and Research with Human Subjects}
    \item[] Question: For crowdsourcing experiments and research with human subjects, does the paper include the full text of instructions given to participants and screenshots, if applicable, as well as details about compensation (if any)? 
    \item[] Answer: \answerNA{} 
    \item[] Justification: The paper does not involve crowdsourcing nor research with human subjects.

\item {\bf Institutional Review Board (IRB) Approvals or Equivalent for Research with Human Subjects}
    \item[] Question: Does the paper describe potential risks incurred by study participants, whether such risks were disclosed to the subjects, and whether Institutional Review Board (IRB) approvals (or an equivalent approval/review based on the requirements of your country or institution) were obtained?
    \item[] Answer: \answerNA{} 
    \item[] Justification: The paper does not involve crowdsourcing nor research with human subjects.

\end{enumerate}

\end{document}